\documentclass[11pt,a4paper]{article}
\usepackage[hyperref]{emnlp2020}
\usepackage{times}
\usepackage{latexsym}
\usepackage{booktabs}
\usepackage{soul}
\usepackage{bm}
\usepackage{microtype}

\aclfinalcopy 

\usepackage{amsmath}
\usepackage{adjustbox}
\usepackage{graphicx}
\usepackage{tipa}
\usepackage{graphicx}

\usepackage{multirow}
\usepackage{multicol,dblfloatfix}

\title{It's All in the Name: A Character Based Approach To Infer Religion}
\author{Rochana Chaturvedi{$^*$} \\
  Keshav Mahavidyalaya \\
  University of Delhi \\
  \texttt{\small{rochana.chaturvedi@gmail.com}} \\ \And
  
  Sugat Chaturvedi{$^*$}   \\
  Economics and Planning Unit \\
  Indian Statistical Institute, Delhi \\
  \texttt{\small{sugat.chaturvedi@gmail.com}} \\}

\date{}

\begin{document}
\maketitle
\def\thefootnote{*}
\footnotetext{The authors have contributed equally to this work.}
\renewcommand{\thefootnote}{\arabic{footnote}}
\begin{abstract}
 Demographic inference from text has received a surge of attention in the field of natural language processing in the last decade. In this paper, we use personal names to infer religion in South Asia\textemdash where religion is a salient social division, and yet, disaggregated data on it remains scarce. Existing work predicts religion using dictionary based method, and therefore, can not classify unseen names. We use character based models which learn character patterns and, therefore, can classify unseen names as well with high accuracy. These models are also much faster and can easily be scaled to large data sets. We improve our classifier by combining the name of an individual with that of their parent/spouse and achieve remarkably high accuracy. Finally, we trace the classification decisions of a convolutional neural network model using layer-wise relevance propagation which can explain the predictions of complex non-linear classifiers and circumvent their purported \emph{black box} nature. We show how character patterns learned by the classifier are rooted in the linguistic origins of names.
\end{abstract}

\section{Introduction}
\label{sec:intro}
Religion is an important marker of identity and is vital in shaping preferences, behavior and economic development outcomes. In India, Hindus are the religious majority comprising 79.8\% of the total population (Census, 2011). Muslims, constituting 14.23\% are the largest minority. Others include Christians (2.30\%), Sikhs (1.72\%), Buddhists (0.70\%) and Jains (0.37\%). Hindus and Muslims have often engaged in violent conflict \citep{varshney2004dataset}. Therefore, the distinction between these two is especially salient. Despite its social and economic relevance, there is a lack of fine-grained data on religion. While such data is collected at the individual level in the Indian Census, it is publicly released as an aggregate only upto the tehsil (sub-district) level. This limits studies on religious demography to only coarse or small scale analyses. This is in contrast to developed countries such as the United States where microdata on race is readily available and has informed studies on racial discrimination as well as residential segregation \citep{cutler1999rise}.

In this paper, we use personal names to predict religion using character based models. Names are often associated with ethnic identity. \citet{bertrand2004emily} demonstrate this for the U.S., where they show evidence of racial discrimination in the labor market by sending fictitious resumes which are randomly assigned African-American or White-sounding names. In South Asia, personal names are especially well known to signify religious identity. \citet{gaikwad2017majority} assign distinctive Hindu or Muslim sounding names to fictitious internal migrants to elicit attitudes of natives towards them in a face to face survey in Mumbai and show that people indeed associate names with religion. 

\begin{figure*}[!htbp]
    \centering
    \includegraphics[trim=0 110 40 95 ,clip,width=.7\textwidth]{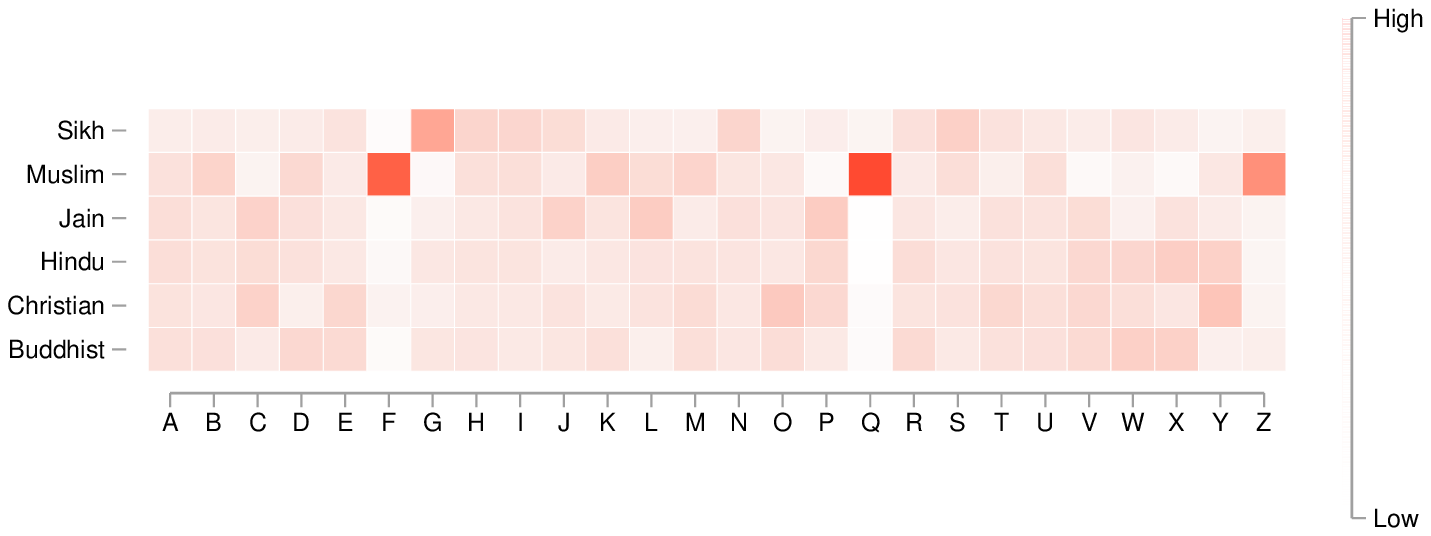}
    \includegraphics[trim=10 120 40 118 ,clip,width=.7\textwidth]{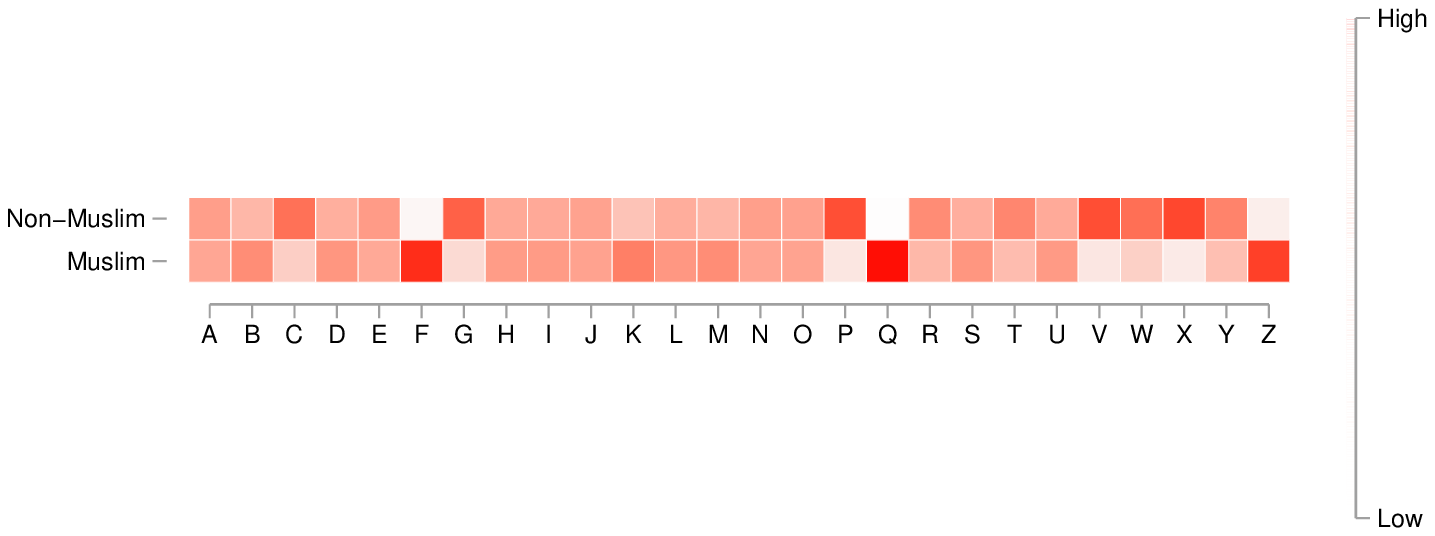}
  \caption{Relative character frequency distributions among names from different religions}
    \label{fig:characters}
\end{figure*}

Name lists are often publicly available at high geographic resolutions. Sources such as electoral rolls, below poverty line (BPL) lists, land records, and beneficiary lists of social security programs such as job cards for Mahatma Gandhi National Rural Employment Guarantee Act (MNREGA), Swachh Bharat Mission (SBM) etc. provide multiple related names for millions of households but do not disclose religion. Our work can be used to compile detailed statistics on religion and uncover evidence of discrimination in the allocation of targeted welfare programs. Similarly, compiling religion from names on social media platforms such as Twitter, which lack demographic attributes, can guide network and sentiment analysis of users across religious groups. In an applied work, \citet{chaturvedi2020importance} use one of our models to find the religion of all the households (over 25 million) in rural Uttar Pradesh, India. They study how gender quotas in local elections interact with preferences of religious groups to affect provision of public goods. 

\subsection{Linguistic Origins of South Asian Names}\label{sec:linguistic}
We also shine light on the distinct orthography of Muslim and non-Muslim names in South Asia. Understanding the differences in linguistic patterns in names belonging to these two classes is particularly interesting from an NLP standpoint as they differ phonologically owing to their linguistic roots. While Classical Arabic is the liturgical language of Islam, Sanskrit is the principal liturgical language of Hinduism. Buddhism, Sikhism and Jainism also have Sanskrit or languages closely related to Sanskrit such as Pali, Punjabi and Magadhi Prakrit as their liturgical languages. Moreover, people who adopt Christianity in India through religious conversions often retain their original names. In line with this, Islamic names in South Asia are derived mostly from Classical Arabic as well as from Persian and Turkish \citep{schimmel1997islamic}. non-Muslim names, on the other hand, are rooted in Sanskrit or Dravidian languages \cite{emeneau1978towards}. These differences manifest in character patterns across names.

Figure \ref{fig:characters} shows raw relative character frequency distributions in our data set. This is obtained for each alphabet A\textendash Z by dividing the average frequency of occurrence of each character by average name length for each religion. We find that Muslim names have a relative abundance of `F', `Q' and `Z' as the phonemes [\textipa{f}], [\textipa{q}] and [\textipa{z}] corresponding to these alphabets are not part of the Sanskrit phonemic inventory. On the other hand, the characters `P', `V' and `X' are rare among Muslim names owing to the absence of the phonemes [\textipa{p}], [\textipa{\textscriptv}] and [\textipa{\textrtails}] in Classical Arabic. Hindu, Sikh, Jain and Buddhist names have similar distributions owing to their common linguistic origins. 

We systematically uncover prominent linguistic patterns by applying layer-wise relevance propagation (LRP) technique proposed by \citet{bach2015pixel}. LRP overcomes the \emph{black box} nature of complex non-linear machine learning models by estimating the extent to which each input contributes towards a particular classification decision. We discuss this in more detail in Section \ref{sec:lrp}.

\subsection{Related Literature}
We contribute to a growing literature in NLP which focuses on demographic inference from names. Early papers in this literature almost exclusively use dictionary based methods to identify ethnicity (see \citealp{mateos2007review} for a detailed review). Others use sub-name features to classify names into nationality, gender and ethnicity using hierarchical decision trees and hidden Markov model \citep{ambekar2009name}, Bayesian inference \citep{chang2010epluribus}, multinomial logistic regression \citep{treeratpituk2012name, torvik2016ethnea} and support vector machine (SVM) \citep{knowles2016demographer}. More recently, deep neural networks (DNN) have been used to infer nationality \citep{lee2017name}, gender and ethnicity \citep{wood2018predicting} using names. \citet{ye2017nationality} and \citet{ye2019secret} learn name embeddings to predict ethnicity, gender and nationality. A vast literature on demographic inference also utilizes social media text \citep{mislove2011understanding, burger2011discriminating, pennacchiotti2011machine, nguyen2013old, ciot2013gender, liu2013s, nguyen2014predicting, volkova2015inferring}.

However, religion inference, which is at least as relevant as ethnicity inference has only received a limited attention. An exception is \citet{susewind2015s}, who uses a string matching algorithm to predict religion based on a given reference list. However, being a dictionary based method, the algorithm suffers from low coverage and can not classify unseen names or names that can not be matched due to spelling variations. We show that simple \emph{bag-of-n-grams} models such as logistic regression (LR) and support vector machines (SVM) combined with term frequency-inverse document frequency (TF-IDF) weights perform at least as well as complex neural models such as convolutional neural networks (CNN), and both outperform the existing work on religion classification.

Much of the applied work on religion in social sciences relies partly or wholly on manual classification of names to infer religion. In particular, \citet{sachar2006social} highlight economic and social disadvantages as well as discrimination faced by Indian Muslims in getting access to credit, by manually classifying religion using personal names. Others use candidate names from the ballot to study the effect of co-religiosity on voting behaviour and Muslim political representation on education and health outcomes of constituents \citep{bhalotra2014religion, heath2015muslim}. Similarly, \citet{field2008segregation} use names in electoral rolls of Ahmedabad to identify religion and study the link between residential segregation and communal violence between Hindus and Muslims during the 2002 Gujarat riots. However, manual classification of names is not feasible for large data sets. Therefore, we hope that our work can contribute towards a richer understanding of economic conditions of various religious groups, discrimination, residential segregation, conflict and the political economy of religion.

\begin{table}[!ht]
\centering
\begin{adjustbox}{max width=.44\textwidth}
\begin{tabular}{lcc}
\hline
 & \textbf{REDS} & \textbf{Rural U.P.} \\ \hline
Observations & 115,180 & 20,000 \\
\# Unique names & 98,853 & 12,348 \\
\# Unique characters & 27 & 27 \\
Average name length & 15.57 & 8.82 \\
Longest name length & 40 & 29\\
\% Buddhist & 0.30 & \textendash \\
\% Christian & 2.46 & \textendash \\
\% Hindu & 84.47 & \textendash \\
\% Jain & 0.42 & \textendash\\
\% Sikh & 3.22 & \textendash \\
\% Muslim & 9.13  & 13.31 \\
\% Non-Muslim & 90.87 & 86.69 \\

\hline
\end{tabular}
\end{adjustbox}
\caption{\label{summary-table}Descriptive statistics}
\end{table}

\section{Data}
\subsection{Data Sources}
\paragraph{REDS} We use Rural Economic \& Demographic Survey (REDS) data collected by the National Council of Applied Economic Research to train our models. The data set is proprietary and not publicly accessible. It constitutes a nationally representative sample of rural households from 17 major Indian states. We use information on the respondent's name, their parent/spouse's name and self-reported religion from more than 115,000 households from a sample of villages surveyed in 2006. We split the REDS data set into three parts\textemdash training, validation and test set in the ratio 70:15:15.

\paragraph{U.P. Rural Households} One possible concern with self-reported religion in the REDS data set could be that some people might not accurately reveal their religion, for example, due to fear of persecution. This might be a source of noise in our data, and we expect that our models would have been even more accurate if there was no misreporting. Keeping this concern in mind and the fact that REDS data contains sensitive information that can not be shared publicly, we use a second test set to further validate our models. Due to lack of public availability of data sets containing a mapping of names to religion, the authors of this paper annotate the religion of 20,000 randomly selected household heads from a data set comprising over 25 million households in rural Uttar Pradesh (U.P.)\textemdash the largest state of India.\footnote{We share the annotated data publicly. These names are in public domain and scraped from \url{https://sbm.gov.in}.} Note that Hindus (comprising 83.66\%) and Muslims (comprising 15.55\%) are the predominant religious groups in rural U.P. and form over 99.2\% of the population. Therefore, the annotators classify the religion as either Non-Muslim (largely comprising Hindus) or Muslim for this data set. The annotations are done independently by the two annotators using the names of household heads as well as the names of their parent/spouse. As discussed earlier, manually annotating religion is in line with the existing literature in the field. The inter-annotator agreement rate is 99.91\% and Kohen's Kappa $\kappa$ is 0.9959 indicating that names strongly reflect religious identity for this sample. The small number of disagreements were resolved on a case-by-case basis in consultation with a Professor of Sanskrit language.
Table \ref{summary-table} shows descriptive statistics for both data sets. The religious composition in our sample closely matches the respective rural population shares for all religious groups.

\subsection{Pre-processing}
We pre-process the data by removing special characters, numbers and extra spaces and retain only alphabetic characters. We then convert all the names to upper case. We enclose each part of name using special start of word and end of word characters. For combining person names with those of their relative, we use special separator character. Each religion is treated as a distinct class.

\section{Models}
We make predictions using single names as well as two names (i.e., primary name and parent/spouse name) in each household. For single name models, we also include the father/spouse's name as a primary name to enrich our training set as it is highly likely that they share the same religion. Since REDS is a nationally representative survey, we do not remove duplicate names to take into account the frequency with which a name occurs within a religion.

\subsection{Baseline: Name2community}
We use a dictionary based classification algorithm name2community proposed by \citet{susewind2015s} as the baseline. It first extracts each name part (i.e. the first name, last name etc.) and counts their frequency in each religious class in a given reference list using spelling (S) and pronunciation (P) matches based on the fuzzy Indic Soundex algorithm. The algorithm then computes a certainty index I for each name part X for each community Y using the formula given below and multiplies it by ``quality factors'' based on spelling and pronunciation $q_S$ and $q_P$ defined as the percentage of unambiguous name parts in the reference list:
\begin{equation*}
    \frac{I(X \in Y)}{q_S\cdot q_P} = \left(1 - \frac{S_{X} - S_{X,Y}}{S_X}\times\frac{P_{X} - P_{X,Y}}{P_X}\right)
\end{equation*}

These indices are then aggregated over all name parts to get the certainty index for the entire name N belonging to a certain community as follows:

\begin{equation*}
    I(N \in Y) = 1 - \left(\prod_{X} \frac{E_X-I(X\in Y)}{E_X} \right)
\end{equation*}

Where $E_X$ refers to the total number of matches of a given name part in the reference list. Finally, each name is assigned the religion with the highest certainty index.\footnote{The implementation of the algorithm is available at \hyperlink{https://github.com/raphael-susewind/name2community}{https://github.com/raphael-susewind/name2community}.}

\subsection{Bag-of-n-grams Models}
We convert each name (or document) to its n-gram feature representation using term frequency-inverse document frequency (TF-IDF). In our case this improves performance over the feature construction procedure described in \citet{knowles2016demographer} who separately extract handcrafted features for a name.\footnote{These include up to four character long prefix and suffix, the entire name, and categorical variables representing whether a particular name begins or ends with a vowel.} TF-IDF captures the importance of each character n-gram (or token) in a document normalized by its importance in the entire corpus without taking into account its relative position in the document. For each token t and document d, the TF-IDF score is calculated as follows:

\begin{equation*}
    TF-IDF(t,d) = TF(t,d)\times IDF(t)
\end{equation*}

\begin{equation*}
    TF(t,d) = \frac{N_{t,d}}{N_d} \quad \text{and,}
\end{equation*}
\begin{equation*}
 IDF(t) = ln\frac{1+n}{1+ DF(t)} + 1
\end{equation*}

$N_{t,d}$ is the number of occurrences of token $t$ in document $d$; $N_d$ is length of document $d$; $DF(t)$ is the number of documents containing token $t$; and $n$ is the count of documents in the corpus.\footnote{The TF-IDF vectors for each document are further normalized to have Euclidean norm 1.} 

We use linear SVM and Logistic Regression (LR) classifiers vectors with l2 regularization. Since the classes are highly imbalanced, we use balanced class weights.

\subsection{Convolutional Neural Network}
Well known for their ability to extract local feature correlations using far fewer parameters and highly parallelizable architectures, convolutional neural networks (CNN) \citep{lecun1990handwritten} have found much appeal in NLP tasks \citep{collobert2008unified}. \citet{zhang2015character} make use of the CNN architecture using character level input data.

In our implementation, we first apply zero padding on names so that all the names have equal length which is the maximum name length in our data after adding the special characters. Thereafter, each character is converted to a 29 or 30 dimensional embedding vector. We use a single CNN layer with max pooling over time. We use kernel sizes ranging from 1\textendash7, with 50\textendash305 filters. We get best results with tanh or ELU activation in CNN layer. The name representation so obtained is finally passed through a fully connected layer with sigmoid activation. This gives us the probabilities of names belonging to each class. The model is trained to minimize binary cross-entropy loss with balanced class weights. We implement the model using Keras deep learning framework \citep{chollet2015keras}. The Kernel weights are randomly initialized using He uniform or Glorot Uniform distributions \citep{he2015delving}. To prevent overfitting, we use dropout rates of 0.01\textendash0.02 on embedding layer and 0.2 after the CNN layer \citep{hinton2012improving}. We use Nadam optimizer \citep{dozat2016incorporating} and reduce learning rate by a factor of 0.5 if our validation loss does not improve for 2 or 3 epochs. We train the models for 80 epochs using mini-batch size of 512 and save the model with the least validation loss. Table \ref{hyperparameters} and \ref{hyperparameterrange}, Appendix \ref{sec:appendix} list the hyperparameters and the range over which we conduct the hyperparameter search.

We find that the use of more complex architectures such as stacked CNN or dilated CNN to learn skip-gram features did not result in any improvement for our task. We also find that Long short-term memory \citep{hochreiter1997long} and CNN-LSTM \citep{kim2016character} models don't result in further improvement.


\begin{table*}[ht!]
\centering
\begin{adjustbox}{max width=\textwidth}
\begin{tabular}{l|c|cc|cc|cc|cc|cc|cc}
\hline
\multirow{2}{*}{\textbf{Models}}& \multirow{2}{*}{\bm{$F_1$}} & \multicolumn{2}{c|}{\textbf{Buddhist}} & \multicolumn{2}{c|}{\textbf{Christian}} &  \multicolumn{2}{c|}{\textbf{Hindu}} & \multicolumn{2}{c|}{\textbf{Jain}}&
\multicolumn{2}{c|}{\textbf{Muslim}}&
\multicolumn{2}{c}{\textbf{Sikh}}\\
\cline{3-14}
& & \textbf{P} & \textbf{R} & \textbf{P} & \textbf{R}&
   \textbf{P} & \textbf{R} & \textbf{P} & \textbf{R} & \textbf{P} & \textbf{R} & \textbf{P} & \textbf{R}\\
 \hline
 \emph{Panel A: Single Name} & & & & & & & & & & & &\\
\multirow{2}{*}{Name2community\bm{$^*$}} & \multirow{2}{*}{44.14} & 0.00 & 0.00 & 26.77 & 44.62 & 92.66 & 95.27 & 26.19 & 26.83 & 90.36 & 87.84 & 43.40 & 14.62  \\
& & (7.17) & (4.24) & (1.58) & (1.78) & (0.30) & (0.27) & (4.28) & (3.80) & (0.89) & (0.77) & (2.20) & (1.12) \\
\multirow{2}{*}{Logistic Regression} & \multirow{2}{*}{71.60} & 64.29 & \textbf{73.97} & 64.04 & \textbf{69.36} & \textbf{97.95} & 94.94 & 28.92 & \textbf{46.15} & 93.17 & \textbf{93.82} & 55.23 & \textbf{91.11} \\
& & (2.28) & (2.75) & (0.98) & (1.15) & (0.18) & (0.19) & (2.29) & (0.33) & (0.53) & (0.59) & (0.69) & (1.00) \\
\multirow{2}{*}{SVM} & \multirow{2}{*}{\textbf{76.90}} & \textbf{89.09} & 67.12 & \textbf{81.27} & 63.90 & 97.39 & \textbf{97.39} & 56.41 & 42.31 & \textbf{95.13} & 93.25 & \textbf{65.43} & 83.48 \\
  & & (2.65) & (2.31) & (1.08) & (0.96) & (0.16) & (0.16) & (3.15) & (2.74) & (0.50) & (0.50) & (0.74) & (0.84) \\
\multirow{2}{*}{CNN} & \multirow{2}{*}{67.88} & 68.25 & 58.90 & 62.04 & 52.02 & 96.64 & 94.23 & \textbf{69.57} & 30.77 & 86.35 & 89.04 & 48.35 & 85.30 \\
 & & (3.03) & (3.02) & (1.28) & (1.26) & (0.20) & (0.21) & (5.02) & (3.58) & (0.60) & (0.65) & (0.77) & (1.10) \\
\hline 
\hline
\emph{Panel B: Concatenated} & & & & & & & & & & & &\\
\multirow{2}{*}{Name2community\bm$^\dagger$} & \multirow{2}{*}{42.42} & 0.00 & 0.00 & 34.19 & 57.14 & 92.67 & 95.55 & 21.21 & 33.33 & 90.58 & 89.53 & 5.32 & 0.97 \\
& & (9.11) & (3.50) & (1.46) & (1.55) & (0.27) & (0.23) & (3.36) & (3.46) & (0.82) & (0.67) & (2.82) & (0.99) \\
\multirow{2}{*}{Logistic Regression} & \multirow{2}{*}{82.76} & 76.00 & 78.08 & 77.18 & \textbf{75.53} & \textbf{98.49} & 97.75 & 75.00 & 57.69 & 95.94 & 96.24 & 76.11 & \textbf{93.10} \\
& & (1.95) & (2.02) & (0.83) & (0.84) & (0.14) & (0.14) & (2.68) & (2.40) & (0.43) & (0.44) & (0.65) & (0.74) \\
\multirow{2}{*}{SVM} & \multirow{2}{*}{\textbf{82.88}} & \textbf{92.86} & 71.23 & \textbf{88.99} & 69.12 & 97.98 & \textbf{98.83} & \textbf{78.12} & 48.08 & \textbf{96.54} & 95.99 & \textbf{83.69} & 85.66 \\
& & (2.19) & (1.82) & (0.91) & (0.76) & (0.14) & (0.13) & (2.90) & (2.15) & (0.41) & (0.39) & (0.69) & (0.66) \\
\multirow{2}{*}{CNN} & \multirow{2}{*}{78.92} & 69.77 & \textbf{82.19} & 69.95 & 70.78 & 98.27 & 97.06 & 44.16 & \textbf{65.38} & 93.92 & \textbf{96.37} & 76.16 & 89.29 \\
& & (1.98) & (2.24) & (0.89) & (0.93) & (0.15) & (0.16) & (2.09) & (2.66) & (0.46) & (0.48) & (0.72) & (0.82) \\
\hline
Observations & 17,207 & \multicolumn{2}{c|}{73} & \multicolumn{2}{c|}{421} & \multicolumn{2}{c|}{14,540} & \multicolumn{2}{c|}{52} & \multicolumn{2}{c|}{1,570} & \multicolumn{2}{c}{551}\\
\hline
\end{tabular}
\end{adjustbox}
\caption{\label{evaluation_test_all} Results for REDS test set. Standard errors reported in parentheses. \bm{$^*$}coverage = 58.35\%; \bm$^\dagger$coverage = 67.34\%. Results in the table only represent the observations classified unambiguously.}
\end{table*}

\section{Results}
\label{sec:results}

\begin{table}[ht!]
\centering
\begin{adjustbox}{max width=.48\textwidth}
\begin{tabular}{l|c|cc|cc}
\hline
\multirow{2}{*}{\textbf{Models}}& \multirow{2}{*}{\bm{$F_1$}} &
\multicolumn{2}{c|}{\textbf{Muslim}} &  \multicolumn{2}{c}{\textbf{Non-Muslim}}\\
\cline{3-6}
& & \textbf{P} & \textbf{R} & \textbf{P} & \textbf{R}\\
 \hline
 \emph{Panel A: Single Name} & & & &\\
\multirow{2}{*}{Name2community\bm{$^*$}} & \multirow{2}{*}{93.11} & 92.58 & 83.20 & 97.98 & 99.19 \\
& & (0.47) & (0.42) & (0.15) & (0.15) \\
\multirow{2}{*}{Logistic Regression} & \multirow{2}{*}{89.95} & 82.79 & \textbf{82.35} & \textbf{97.29} & 97.37 \\
& & (0.40) & (0.40) & (0.16) & (0.15) \\
\multirow{2}{*}{SVM} & \multirow{2}{*}{\textbf{91.47}} & \textbf{90.13} & 80.59 & 97.07 & \textbf{98.64} \\
& & (0.39) & (0.35) & (0.14) & (0.14) \\
\multirow{2}{*}{CNN} & \multirow{2}{*}{85.67} & 71.08 & 80.29 & 96.91 & 94.98 \\
& & (0.43) & (0.48) & (0.18) & (0.19) \\
\hline 
\hline
\emph{Panel B: Concatenated} & & & & & \\
\multirow{2}{*}{Name2community\bm$^\dagger$} & \multirow{2}{*}{91.62} & 85.92 & 84.84 & 97.79 & 91.62 \\
& & (0.43) & (0.42) & (0.16) & (0.16) \\
\multirow{2}{*}{Logistic Regression} & \multirow{2}{*}{95.99} & 96.12 & 90.12 & 98.50 & 99.44 \\
& & (0.27) & (0.25) & (0.10) & (0.10) \\
\multirow{2}{*}{SVM} & \multirow{2}{*}{\textbf{96.10}} & \textbf{98.21} & 88.66 & 98.28 & \textbf{99.75} \\
& & (0.27) & (0.24) & (0.10) & (0.09) \\
\multirow{2}{*}{CNN} & \multirow{2}{*}{94.34} & 87.54 & \textbf{93.09} & \textbf{98.93} & 97.96 \\
& & (0.29) & (0.31) & (0.12) & (0.12) \\
\hline
Observations & 20,000 & \multicolumn{2}{c|}{2,663} & \multicolumn{2}{c}{17,337}\\
\hline
\end{tabular}
\end{adjustbox}
\caption{\label{evaluation_test_sbm} Results for U.P. Rural Households test set. Standard errors reported in parentheses. \bm{$^*$}coverage = 57.26\%; \bm$^\dagger$coverage = 74.74\%. Results in the table only represent the observations classified unambiguously.}
\end{table}

We report the results of our experiments in Table \ref{evaluation_test_all} for the REDS test set and in Table \ref{evaluation_test_sbm} for the U.P. Rural Households data set. For the second test set, we use a two-way classification of religion as Muslim or non-Muslim based on our multi-class models. Panel A shows the results when predicting religion using only a single name. The name2community algorithm \citep{susewind2015s} is only able to classify less than 60\% of observations in both the test sets. The scores in the table are reported only for the observations classified unambiguously. We also experiment by assigning\textemdash as a tie-breaking rule\textemdash the majority religion to the observations which could not be classified using name2community. This reduces the macro-average $F_1$ score to 38\% and 75\% for the two data sets respectively. 

On the other hand, character based models perform exceptionally well and have 100\% coverage. Even among names that could be assigned a class using name2community, the overall accuracy is significantly higher for LR, SVM, as well as the CNN models for the REDS test set. In fact, along with 100\% coverage, SVM has higher precision and recall than name2community for all the classes. The differences in recall are statistically significant at 5\% level for the Buddhist class, and at 1\% level for all the other classes. Further, in our experiments name2community was orders of magnitude slower than the character based models. It could only predict 0.4 names/second. In contrast, other models predicted 50,000\textendash 500,000 names/second. Therefore, predicting religion for the entire REDS test set comprising 17,207 names took nearly 12 hours using name2community for single name and over 19 hours with concatenated names. On the other hand, all the character based models only took 0.035\textendash0.4 seconds to predict all the names. This makes name2community less scalable to large data sets that may comprise millions of observations and further limits its viability as a general religion classifier for South Asia.

Comparing the bag-of-n-grams models with CNN reveals that the accuracy of LR and SVM is significantly higher compared to the CNN model at 1\% level of significance. They also have higher macro-average $F_1$ scores. This implies that the classes can be separated linearly in the very high-dimensional character n-gram space. The results improve\textemdash especially for the CNN model\textemdash when we enrich our data by concatenating individual names with their parent/spouse's name as shown in panel B. This improvement is primarily driven by better identification of minority names which comprise a smaller number of observations in our sample, and hence can benefit from richer data. The $F_1$ score for the CNN model is now closer, but still much lower than LR and SVM. The overall accuracy is also lower for the CNN model and the difference is statistically significant at 1\% level. However, the recall for Muslim names is slightly better for the CNN model resulting in more balanced predictions. Looking at panel B, we also note that the coverage for name2community increases by 9 percentage points for REDS data, and 17 percentage points for the rural U.P. data using the concatenated model. However, this is accompanied by a decline in the macro-average $F_1$ score. This shows that there are only limited gains from providing a richer data to name2community.\footnote{We report validation set results in Table \ref{evaluation_val_all}, Appendix \ref{section:eval}.}

For all our models, most of the incorrect predictions for Buddhist, Christian, Jain, and Sikh classes end up being classified as Hindu class. This not only reaffirms the hypothesis that the common linguistic origins of Buddhists, Hindus, Jains and Sikhs are reflected in personal names, but also suggests that Christian converts in India often retain their original Hindu names. To illustrate, we show confusion matrices for SVM models in Table \ref{confusion_matrix}, Appendix Section \ref{sec:confusion}.


\begin{table*}[t!]
\centering
\begin{adjustbox}{max width=\textwidth}
\begin{tabular}{ccc}
\hline
\textbf{S.no.} & \textbf{True Muslims} & \textbf{True non-Muslims}\\
\hline
& & \\
1. & \raisebox{-.2\totalheight}{\includegraphics[width=\textwidth]{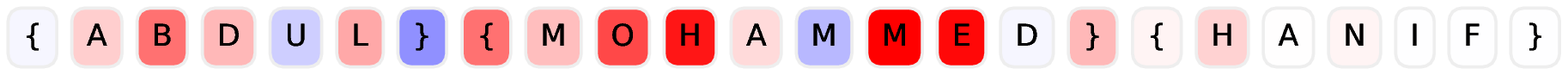}} & \raisebox{-.2\totalheight}{\includegraphics[width=\textwidth]{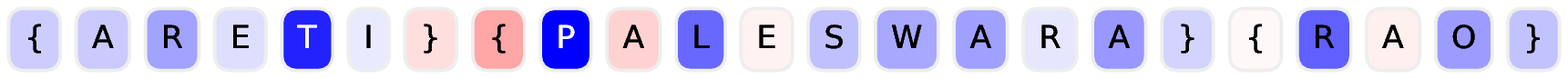}} \\
2. & \raisebox{-.2\totalheight}{\includegraphics[width=\textwidth]{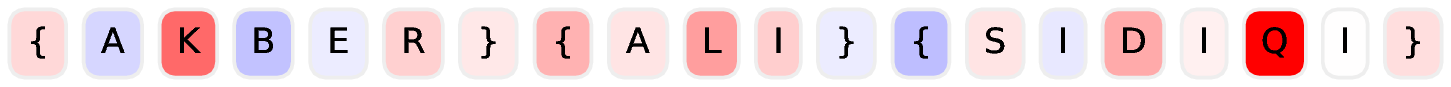}} & \raisebox{-.2\totalheight}{\includegraphics[width=\textwidth]{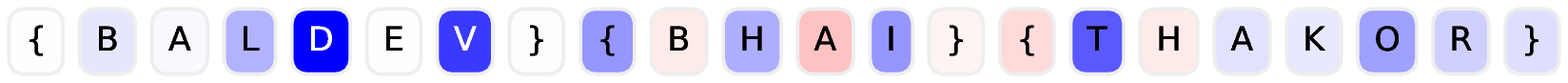}} \\
3. & \raisebox{-.2\totalheight}{\includegraphics[width=\textwidth]{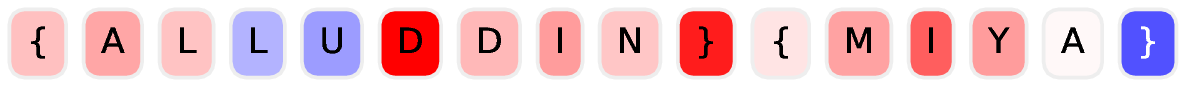}} & \raisebox{-.2\totalheight}{\includegraphics[width=\textwidth]{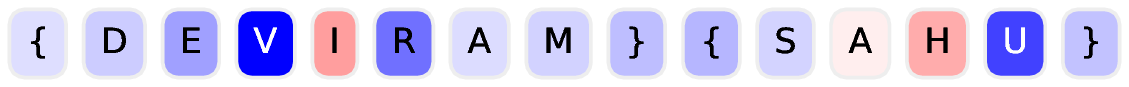}} \\
4. & \raisebox{-.2\totalheight}{\includegraphics[width=\textwidth]{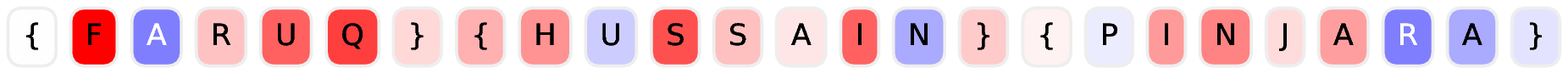}} & \raisebox{-.2\totalheight}{\includegraphics[width=\textwidth]{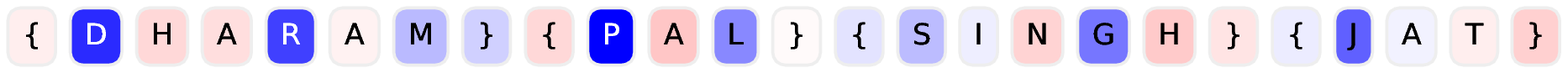}} \\
5. & \raisebox{-.2\totalheight}{\includegraphics[width=\textwidth]{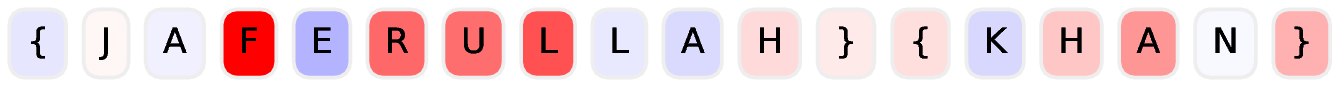}} & \raisebox{-.2\totalheight}{\includegraphics[width=\textwidth]{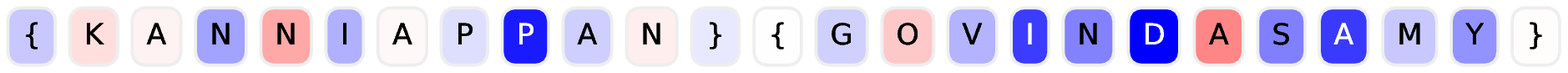}} \\
6. & \raisebox{-.2\totalheight}{\includegraphics[width=\textwidth]{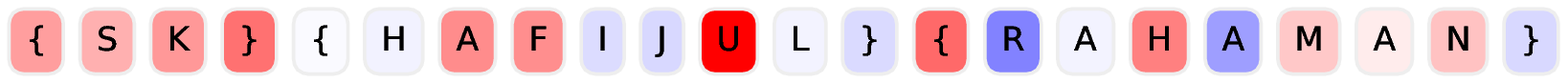}} & \raisebox{-.2\totalheight}{\includegraphics[width=\textwidth]{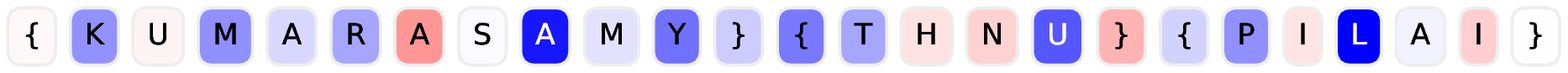}} \\
7. & \raisebox{-.2\totalheight}{\includegraphics[width=\textwidth]{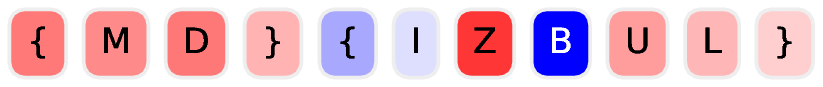}} & \raisebox{-.2\totalheight}{\includegraphics[width=\textwidth]{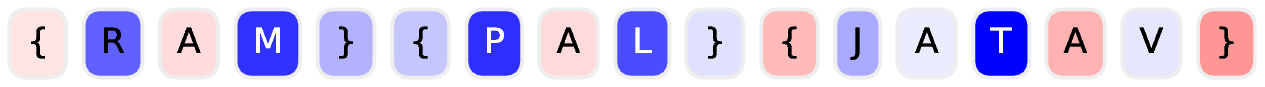}} \\
8. & \raisebox{-.2\totalheight}{\includegraphics[width=\textwidth]{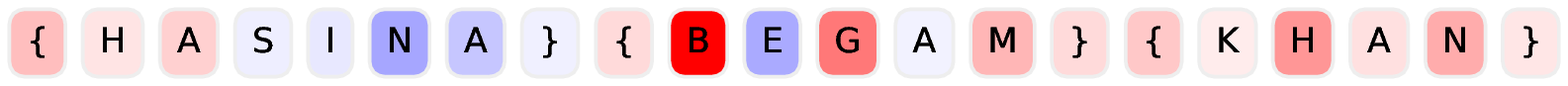}} & \raisebox{-.2\totalheight}{\includegraphics[width=\textwidth]{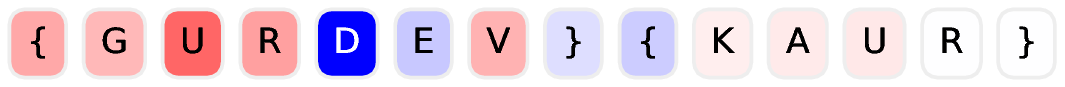}} \\
9. & \raisebox{-.2\totalheight}{\includegraphics[width=\textwidth]{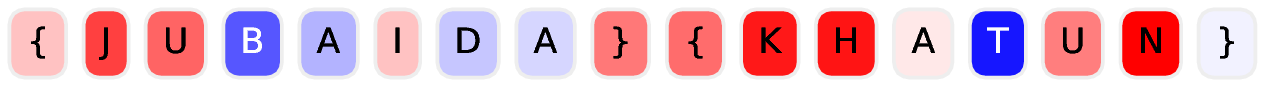}} & \raisebox{-.2\totalheight}{\includegraphics[width=\textwidth]{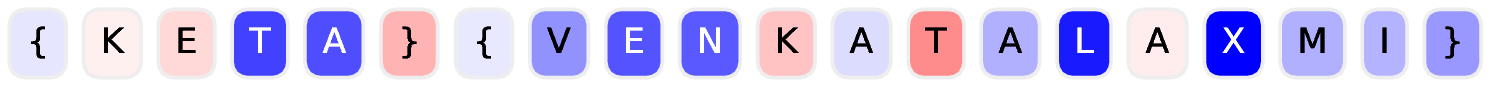}} \\
10. & \raisebox{-.2\totalheight}{\includegraphics[width=\textwidth]{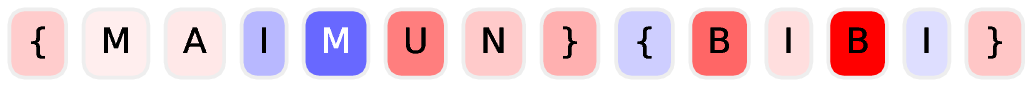}} & \raisebox{-.2\totalheight}{\includegraphics[width=\textwidth]{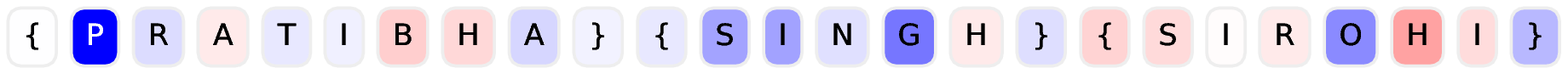}} \\
& & \\
\hline
\end{tabular}
\end{adjustbox}
\caption{\label{lrp}LRP heatmaps of distinctive test set names. Muslim relevance is red, non-Muslim relevance is blue.}
\end{table*}

\section{Explaining Predictions Using Layer-wise Relevance Propagation}\label{sec:lrp}
In this section, we focus on which character patterns help in differentiating Muslim from non-Muslim names in India. As discussed before, these two classes have different linguistic roots. Moreover, our observations from the previous section also give us a compelling reason to focus on differences between these two classes. We train a set of binary classifiers for these two classes for this exercise. The model details and results are reported in Appendix Section \ref{sec:twoway}. We apply layer-wise relevance propagation (LRP) on REDS test set to identify underlying linguistic differences between names. Explainability of a model is also important to assess its validity and generalizability, and to foster trust in its predictions.\footnote{\citet{lapuschkin2019unmasking} find that a Fisher Vectors based model spuriously achieved state-of-the-art accuracy on an image classification task by misusing source tags that often occurred in lower left corners of horse images instead of learning about the actual horse in the image.}

LRP uses backward propagation to redistribute the output ``classification score'' obtained before applying the softmax function to the input layer. This explains the relevance of each input towards the classification decision of the model. The backward propagation satisfies a layer-wise conservation principle, i.e., the sum of relevances for each layer (including the input) equal the classification score. The relevance of each character $c$ is then calculated from the input vectors by simply summing the input relevance scores $R_{c,d}$ over all the dimensions $d$ of the vector:
\begin{equation*}
    R_w = \sum_{d}R_{c,d}
\end{equation*}

As opposed to the gradient based Sensitivity Analysis (SA), LRP directly decomposes the actual value $f(x)$ returned by the machine learning model $f$ for an input $x$, and thus, can identify characters having positive as well as negative relevance for each class. Therefore, LRP directly answers what character patterns make a name Muslim, instead of what change in character patterns will make a name more or less Muslim.

\begin{table*}[!htbp]
\centering
\begin{adjustbox}{max width=.6\textwidth}
\begin{tabular}{cccccc}
\hline
\multicolumn{2}{c}{\textbf{Unigram}} & \multicolumn{2}{c}{\textbf{Bigram}} & \multicolumn{2}{c}{\textbf{Trigram}}\\
Muslim & Non-Muslim & Muslim & Non-Muslim & Muslim & Non-Muslim \\\hline
 F & X & F\} & PR & F\}\{ & PRA \\
 Q & V & IF & IV & \{SK & DEV\\
 Z & P & AF & GW & SAB & \{PR\\
 B & W & FI & EV & SK\} & GHO\\
 H & G & FA & VV & DDI & SIV\\
 J & C & B\} & SW & AB\} & EV\}\\
 U & T & KH & MP & KH\} & EGH\\
 \textendash & Y & FU & EP & FAR & DEY\\
 \textendash & R & DD & LD & BI\} & PAL \\
 \textendash & O & ZA & V\} & ED\} & VVA \\
\hline
\end{tabular}
\end{adjustbox}
\caption{\label{lrpchar} Ten most important unigrams, bigrams and trigrams identified by LRP using CNN classifier for each class conditional on the n-grams not being rare, i.e., occurring at least 25 times in the test set.}
\end{table*}

LRP has been variously applied to explain classification decisions of DNN on image \citep{samek2016evaluating, binder2016layer}, speech \citep{becker2018interpreting}, and video data \citep{anders2019understanding}. It has also proven its usefulness in the medical domain for detecting brain activity \citep{sturm2016interpretable, thomas2019analyzing} and cancer biomarkers \citep{hagele2020resolving, binder2018towards}. \citet{arras2017relevant} and \citet{arras2017explaining} apply LRP to text data for CNN and bi-LSTM models respectively and show that it provides better explanations than SA. They also show that though both SVM and CNN models performed comparably in terms of classification accuracy, the explanations from the CNN model were more human interpretable. This is because CNN takes into account the context in which tokens appear as opposed to SVM which only considers n-gram frequencies. Therefore, we study the decisions of our CNN model using the LRP implementation of \citet{ancona2017towards}.\footnote{\url{https://github.com/marcoancona/DeepExplain}} 

In Table \ref{lrp}, we show LRP heatmaps showing classification decisions of CNN model on distinctive names from REDS test set. Characters with positive relevance scores with respect to Muslim class are labelled red, while those with negative relevance scores are labelled blue, i.e., they have positive relevance for the non-Muslim class. The left panel shows examples of correctly classified Muslim names. The LRP relevance scores are able to identify phonemes characteristic in Classical Arabic such as `F' (column 1, example 4, 5, 6), `Q' (column 1, example 2, 4), `Z' (column 1, example 7) as well as `KH' (column 1, example 9). Meaningful prefixes such as `ABD' (column 1, example 1) meaning ``slave" (usually followed by `UL' meaning ``of" which is usually followed by the name of God) and suffixes such as `UDDIN' (column 1, example 3) meaning ``(of) the religion/faith/creed" which are highly characteristic of Arabic names are also detected as relevant for the Muslim class. The right panel shows correctly classified non-Muslim names. The phonemes such as `P' (column 2, example 1), `V' (column 2, example 2) and `X' (column 2, example 9) are highly relevant for the non-Muslim class.

For linear models, the relevance values are computed per character n-gram feature implying that the relevance of a character is independent of its context. This is not the case for neural models where relevance scores of characters n-grams depend on other characters in the name. We confirm this as the neutral character `D' is highly relevant to the Muslim class when it is a part of `UDDIN', while its relevance score flips and it becomes highly relevant to non-Muslim class when it forms the word `DEV' (meaning god in Sanskrit) (column 2, examples 2, 3 and 8).\footnote{We see similar contrast for (1) `L' in `UL' (meaning `of' in Arabic) vs. `PAL' and (2) `M' in `MOHAMMED' vs. `RAM' (the name of a Hindu deity).} We notice that the relevance of the characters `\{' and `\}' signifying the beginning and end of a name part respectively is also modulated by the character sequences following and preceding them respectively.

In Table \ref{lrpchar} we show the ten most relevant character unigrams, bigrams and trigrams for both Muslim and non-Muslim target classes using LRP on test set names, and average the relevance scores for each n-gram.\footnote{The layer-wise conservation principle described earlier allows us to simply sum up relevance scores of constituent characters to calculate the relevance score for each n-gram.} We note that the linguistic differences noted in Section \ref{sec:linguistic} are indeed captured by the CNN model. Unigrams `F', `Q' and `Z' are most predictive of the Muslim class, whereas `X', `V' and `P' are most relevant for the non-Muslim class. As expected, we also find the bigram `KH' corresponding to phoneme [\textipa{x}] is highly relevant for the Muslim class. A look at the most relevant bigrams for the Muslim class also shows the importance of character position in a name part. We find that `F\}' and `B\}' are highly relevant to the Muslim class implying that the characters `B' or `F' at the end of a name part are a distinguishing feature of Muslim names. On the other hand, the bigram `PR' is a distinguishing feature of the non-Muslim class, especially when it occurs at the beginning of a name part, denoted by the trigram `\{PR'. This is meaningful as `PR' is a Sanskrit prefix which when added to an adjective or a noun accentuates its quality. Similarly, the bigram `VV' has positive relevance for non-Muslim class as it forms part of the Dravidian honorific suffix `AVVA' added to female names. The trigram `DDI' is considered highly relevant by our model and forms part of the suffix `UDDIN' in Arabic names. While a comprehensive analysis of naming patterns is out of the scope of this paper, these examples illustrate that LRP relevances are very reliable at finding meaningful character n-grams that distinguish the two classes and highlight the linguistic differences depicted by the names.

\begin{figure*}[!t]
    \centering
    \begin{minipage}[htbp!]{.49\textwidth}
    \includegraphics[width=\textwidth]{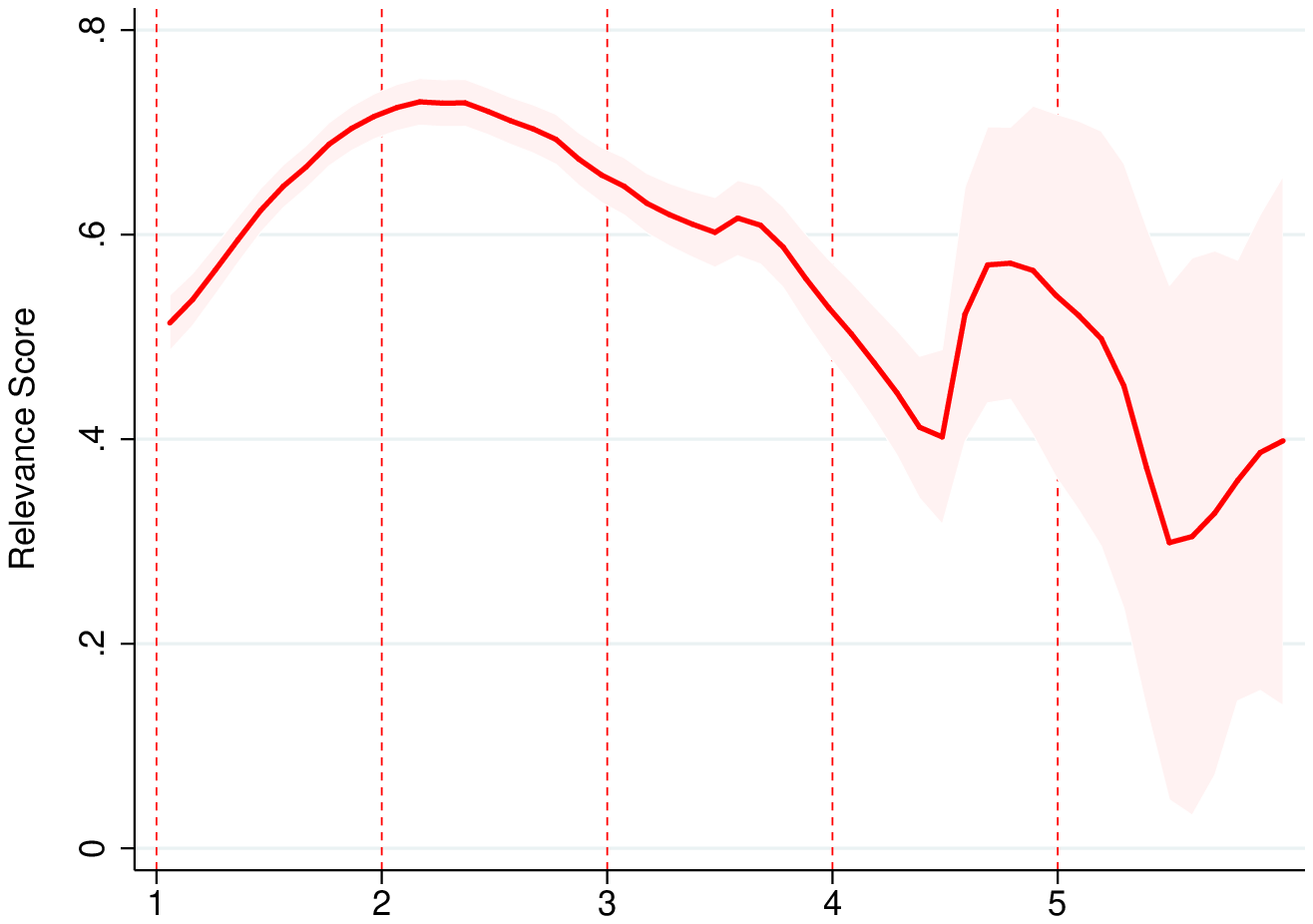}
    \end{minipage}
    \begin{minipage}[htbp!]{.49\textwidth}
    \includegraphics[width=\textwidth]{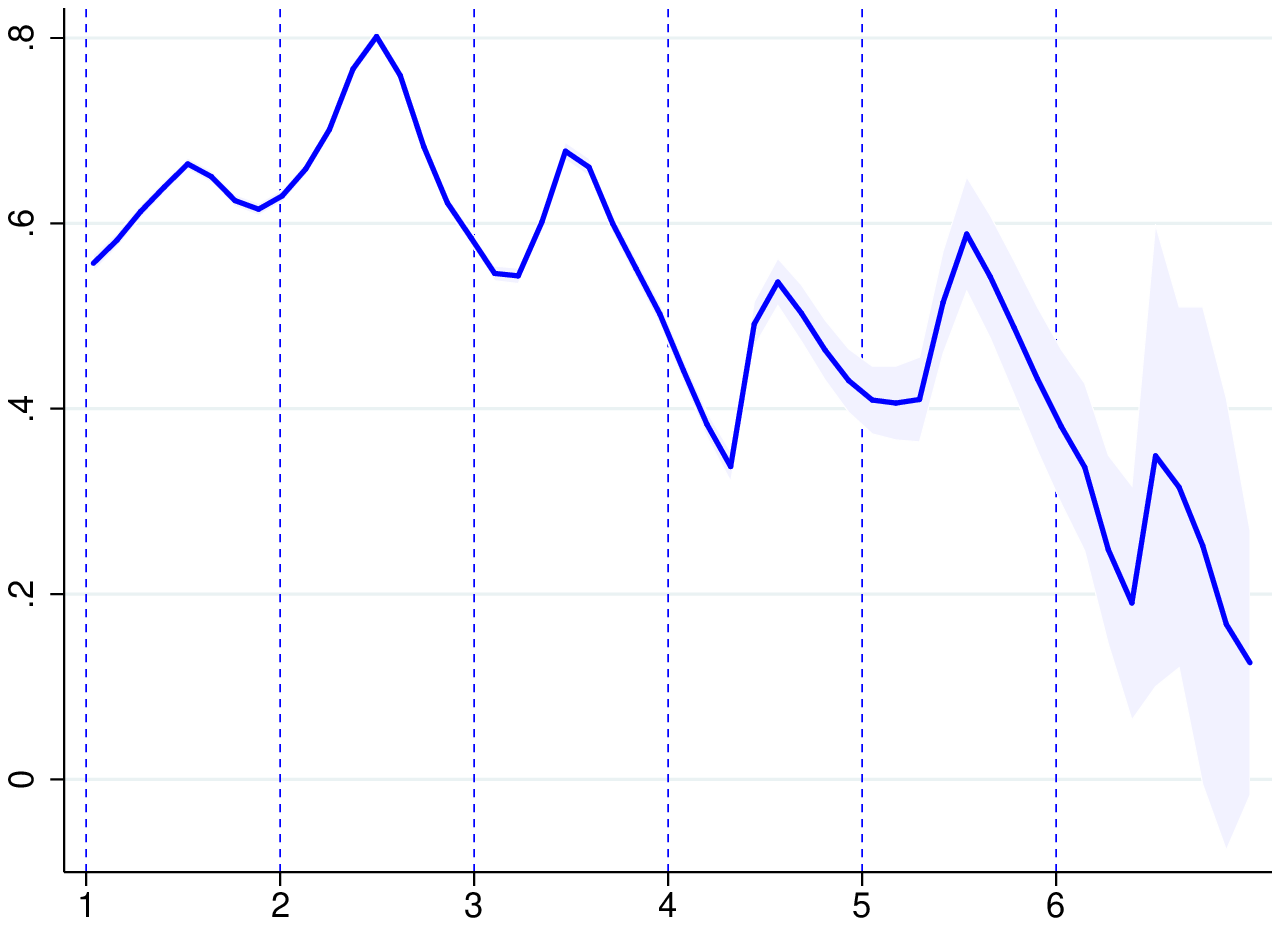}
    \end{minipage}
  \caption{Character relevance distribution over name length using LRP on test sample. Left panel shows Muslim relevance for True Muslims in the sample and right panel shows Non-Muslim relevance for true Non-Muslims. The x-axis represents $i^{th}$ name part. The shaded region denotes 95\% confidence interval.}
    \label{fig:pos}
\end{figure*}

To understand which part in a name contributes the most to predictions for a certain class, we perform LRP on all the correctly classified names in the test set. We then divide each observation into name parts (i.e. first name, last name etc.), and map relevance scores to normalized length of name part such that the maximum length of a name part is 1. We use absolute values of character relevance scores. Figure \ref{fig:pos} shows local polynomial plot of relevance scores over name length. We find that for correctly classified Muslims (left panel), major part of the relevance is attributed to the end of the first name part and to the beginning of the second name part by the model. For correctly classified Hindus (right), however, the highest relevance for the Hindu class is concentrated in the middle of each name part\textemdash especially the second name part. In both the panels, we find that the latter name parts are considered less important by our classifier.


\section{Ethical Implications}
Religion is a sensitive issue and religious minorities face unfair treatment in many parts of the world. This is plausibly true for India as well where religious conflict is common and there is evidence of religious discrimination, for instance, in urban housing rental market \citep{thorat2015urban} and labor market \citep{thorat2007legacy}. Therefore, it is of utmost importance to include a discussion on ethical implications of our work. 

A potential risk that comes with any new technology is that it may be used by nefarious actors. In our case, one could argue that the inferences made by our models could exacerbate religious discrimination and violence. This might also pose a risk to people incorrectly classified among the targeted group. However, as discussed earlier, religious connotations of names are well known in India. This means that at the individual level, for instance, while making hiring decisions or renting property, employers and landlords do not require an algorithm to identify religion. Likewise, local rioters can easily use publicly available name-lists to manually identify religion with near perfect accuracy. Moreover, government actors already have access to detailed information on religion through the Indian census and other sources. 

On the other hand, by making the data more accessible to researchers, we expect our work will help highlight any discrimination and deprivations experienced by religious minorities. This could, in turn, help formulate policies to protect the vulnerable groups and thereby reduce religious discrimination. Therefore, we expect that the benefits of such a technology should outweigh its potential costs. We caution that names may not always have a strict correspondence with religion. Therefore, our classification exercise should be interpreted as a probabilistic rather than a rigid mapping.

\section{Conclusion}
In this paper, we infer religion from South Asian names using self-reported survey data. Whereas dictionary based methods to classify religion from names suffer from slow speed, low coverage and low accuracy, character based machine learning models perform exceptionally well at this task. The performance of these models is further improved by leveraging two names within each household in our data. Further, our experiments show that complex neural models such as CNN, which require extensive hyperparameter tuning and more computing resources, do not necessarily provide an advantage over simpler LR and SVM models at this task.

We explain the predictions of one of our models using LRP. To the best of our knowledge, we are the first to apply LRP to explain classification decisions of a character-based classifier. We demonstrate that LRP can be used as a tool to detect linguistic differences encapsulated in names of different religions. A machine learning model that infers religion using personal names can easily be scaled to massive data sets. We reflect on the ethical implications of this, and conclude that given the lack of fine-grained data on religion, and abundance of public name lists, this is a viable way to generate data to study religious demography.

\section*{Acknowledgments}
We thank Mithilesh Chaturvedi, Sabyasachi Das, Kanika Mahajan, Sudha Rao, Raphael Susewind as well as anonymous reviewers for helpful feedback.
\bibliographystyle{acl_natbib}
\bibliography{emnlp2020}

\appendix
\onecolumn
\section{Hyperparameters}
\label{sec:appendix}

\begin{table}[ht!]
\centering
\begin{adjustbox}{max width=\textwidth}
\begin{tabular}{lcc}
\hline
\textbf{Parameter} & \textbf{Single Name} & \textbf{Concatenated} \\ \hline
Embedding dimension & 29 & 30 \\
CNN kernel sizes ($k$) & \{1,2,3,4,5,6,7\} & \{1,2,3,4,5,6,7\}\\
CNN filters & \{50,300,305,200,250,200,200\} & \{239,248,100,150,150,250,200\} \\
Dense units & 400 & 200 \\
Dense activation & sigmoid & sigmoid \\
CNN activation & tanh & ELU \\
Embedding dropout & 0.01 & 0.02\\
Dropout & 0.2 & 0.2\\
Loss & binary cross-entropy & binary cross-entropy \\
Batch size & 512 & 512 \\
Optimizer & Nadam & Nadam \\
Kernel initialization & He uniform & Glorot uniform\\
Minimum learning rate & 0.0002 & 0.00027 \\
Batch normalization & True & True \\
Learning rate reduction factor & 0.5 & 0.5 \\
Patience & 2 epochs & 3 epochs \\
epochs & 80 & 60 \\
l2 regularization parameter & 64.49 (LR), 79.53 (SVM)  & 33.39 (LR), 8.47 (SVM)\\
TF-IDF Max n-grams & 12 (LR), 11 (SVM) & 10 (LR, SVM) \\
\hline
\end{tabular}
\end{adjustbox}
\caption{\label{hyperparameters} Hyperparameter choice}
\end{table}

\begin{table*}[!h]
\centering
\begin{adjustbox}{max width=\textwidth}
\begin{tabular}{lc}
\hline
 \textbf{Parameter} & \textbf{Experimental Range} \\ \hline
Embedding dimension & 5-80\\
CNN kernel sizes (k) & 1\textendash7\\
CNN filters & 0\textendash300\\
Dense units & 0\textendash400 \\
Dense activation & ReLU, tanh, sigmoid \\
CNN activation & ELU, ReLU, tanh \\
Embedding dropout & 0\textendash0.25\\
Dropout & 0\textendash0.5\\
Loss & binary cross-entropy, focal-loss\\
Batch size & 32\textendash1024\\
Optimizer & Adam, Nadam, rmsprop \\
Kernel initialization & He uniform, Glorot uniform, He Normal\\
Minimum learning rate & $1\times10^{-5}$ \textendash $1\times 10^{-3}$ \\
Batch normalization & True, False \\
Learning rate reduction factor & 0.5\textendash0.8 \\
Patience & 3\textendash5 epochs \\
epochs & 20\textendash80 \\
l2 regularization parameter & 0\textendash100\\
TF-IDF Max n-grams & 1\textendash12\\

\hline
\end{tabular}
\end{adjustbox}
\caption{\label{hyperparameterrange} Hyperparameter range}
\end{table*}

\clearpage
\section{Confusion Matrices for REDS Test set with SVM Classifier}\label{sec:confusion}

\begin{table}[ht!]
\centering
\begin{adjustbox}{max width=\textwidth}
\begin{tabular}{lll}
\begin{tabular}{|ll|cccccc|c|}
\hline
& & \multicolumn{6}{c|}{\textbf{Predicted}} & \\
& & \textbf{B} & \textbf{C} & \textbf{H} & \textbf{J} & \textbf{M} & \textbf{S} & \textbf{Total}\\
 \hline
\multirow{6}{*}{\textbf{True}} & \textbf{B} & 49 & 0 & 24 & 0 & 0 & 0 & 73 \\
& \textbf{C} & 0 & 269 & 147 & 0 & 5 & 0 & 421 \\
& \textbf{H} & 6 & 59 & 14,160 & 17 & 55 & 243 & 14,540 \\
& \textbf{J} & 0 & 0 & 30 & 22 & 0 & 0 & 52 \\
& \textbf{M} & 0 & 1 & 105 & 0 & 1,464 & 0 & 1,570 \\
& \textbf{S} & 0 & 2 & 74 & 0 & 15 & 460 & 551 \\
\hline
& \textbf{Total} & 55 & 331 & 14,540 & 39 & 1539 & 703 & 17,207 \\
\hline
\end{tabular}
& &
\begin{tabular}{|ll|cccccc|c|}
\hline
& & \multicolumn{6}{c|}{\textbf{Predicted}} & \\
& & \textbf{B} & \textbf{C} & \textbf{H} & \textbf{J} & \textbf{M} & \textbf{S} & \textbf{Total}\\
 \hline
\multirow{6}{*}{\textbf{True}} & \textbf{B} & 52 & 0 & 21 & 0 & 0 & 0 & 73 \\
& \textbf{C} & 0 & 291 & 125 & 0 & 5 & 0 & 421 \\
& \textbf{H} & 4 & 33 & 14,370 & 7 & 37 & 89 & 14,540 \\
& \textbf{J} & 0 & 0 & 27 & 25 & 0 & 0 & 52 \\
& \textbf{M} & 0 & 0 & 60 & 0 & 1,507 & 3 & 1,570 \\
& \textbf{S} & 0 & 3 & 64 & 0 & 12 & 472 & 551 \\
\hline
& \textbf{Total} & 56 & 327 & 14,667 & 32 & 1561 & 564 & 17,207 \\
\hline
\end{tabular}
\end{tabular}
\end{adjustbox}
\caption{\label{confusion_matrix} Confusion matrix for the single name (left) and concatenated (right) SVM model using REDS test set. \\ B = ``Buddhist'', C = ``Christian'', H = ``Hindu'', J = ``Jain'', M = ``Muslim'', S = ``Sikh''}
\end{table}

\section{Validation Set Performance}\label{section:eval}
\begin{table*}[ht!]
\centering
\begin{adjustbox}{max width=\textwidth}
\begin{tabular}{l|c|cc|cc|cc|cc|cc|cc}
\hline
\multirow{2}{*}{\textbf{Models}}& \multirow{2}{*}{\bm{$F_1$}} & \multicolumn{2}{c|}{\textbf{Buddhist}} & \multicolumn{2}{c|}{\textbf{Christian}} &  \multicolumn{2}{c|}{\textbf{Hindu}} & \multicolumn{2}{c|}{\textbf{Jain}}&
\multicolumn{2}{c|}{\textbf{Muslim}}&
\multicolumn{2}{c}{\textbf{Sikh}}\\
\cline{3-14}
& & \textbf{P} & \textbf{R} & \textbf{P} & \textbf{R}&
   \textbf{P} & \textbf{R} & \textbf{P} & \textbf{R} & \textbf{P} & \textbf{R} & \textbf{P} & \textbf{R}\\
 \hline
 \emph{Panel A: Single Name} & & & & & & & & & & & &\\
\multirow{2}{*}{Name2community\bm{$^*$}} & \multirow{2}{*}{43.64} & 0.00 & 0.00 & 28.86 & 47.25 & 92.45 & 95.52 & 22.73 & 37.04 & 89.88 & 87.07 & 39.47 & 9.57 \\
& & (6.59) & (3.60) & (1.62) & (1.75) & (0.30) & (0.26) & (4.21) & (4.54) & (0.92) & (0.76) & (2.62) & (1.09) \\
\multirow{2}{*}{Logistic Regression} & \multirow{2}{*}{70.20} & 56.32 & \textbf{67.12} & 59.37 & \textbf{66.82} & \textbf{97.72} & 94.81 & 34.85 & \textbf{44.23} & 92.69 & \textbf{92.93} & 55.67 & \textbf{90.93} \\
& & (2.30) & (2.80) & (0.99) & (1.17) & (0.18) & (0.20) & (2.64) & (3.32) & (0.54) & (0.60) & (0.72) & (1.02) \\
\multirow{2}{*}{SVM} & \multirow{2}{*}{\textbf{74.33}} & \textbf{81.63} & 54.79 & \textbf{73.91} & 60.43 & 97.03 & \textbf{97.40} & 61.76 & 40.38 & \textbf{95.30} & 91.78 & \textbf{67.61} & 82.58 \\
  & & (2.92) & (2.36) & (1.10) & (0.98) & (0.17) & (0.17) & (3.50) & (2.79) & (0.53) & (0.51) & (0.79) & (0.86) \\
\multirow{2}{*}{CNN} & \multirow{2}{*}{67.27} & 64.91 & 50.68 & 61.74 & 50.47 & 96.54 & 94.46 & \textbf{72.73} & 30.77 & 86.16 & 88.46 & 51.59 & 88.38 \\
 & & (3.20) & (2.98) & (1.30) & (1.24) & (0.20) & (0.21) & (5.15) & (3.53) & (0.60) & (0.64) & (0.79) & (1.08) \\
\hline 
\hline
\emph{Panel B: Concatenated} & & & & & & & & & & & &\\
\multirow{2}{*}{Name2community\bm$^\dagger$} & \multirow{2}{*}{43.56} & 0.00 & 0.00 & 29.93 & 48.66 & 92.41 & 95.52 & 22.73 & 37.04 & 89.85 & 86.94 & 36.36 & 8.51 \\
& & (6.60) & (3.59) & (1.61) & (1.72) & (0.30) & (0.26) & (4.22) & (4.53) & (0.92) & (0.76) & (2.67) & (1.09) \\
\multirow{2}{*}{Logistic Regression} & \multirow{2}{*}{80.24} & 65.12 & \textbf{76.71} & 74.29 & \textbf{74.64} & \textbf{98.33} & 97.45 & 72.97 & 51.92 & 96.46 & \textbf{95.60} & 73.42 & \textbf{92.74} \\
& & (1.90) & (2.13) & (0.85) &  (0.89) & (0.15) & (0.15) & (2.89) & (2.53) & (0.45) & (0.46) & (0.67) & (0.78) \\
\multirow{2}{*}{SVM} & \multirow{2}{*}{\textbf{80.76}} & \textbf{88.68} & 64.38 & \textbf{87.27} & 66.59 & 97.79 & \textbf{98.67} & \textbf{78.57} & 42.31 & \textbf{96.83} & 95.41 & \textbf{82.45} & 87.84 \\
& & (2.33) & (1.87) & (0.95) & (0.78) & (0.14) & (0.13) & (3.21) & (2.22) & (0.43) & (0.40) & (0.70) & (0.68) \\
\multirow{2}{*}{CNN} & \multirow{2}{*}{76.14} & 63.86 & 72.60 & 63.96 & 68.96 & 97.99 & 96.62 & 41.77 & \textbf{63.46} & 93.65 & 94.96 & 73.76 & 88.75 \\
& & (2.13) & (2.40) & (0.91) & (1.00) & (0.16) & (0.17) & (2.18) & (2.84) & (0.49) & (0.52) & (0.75) & (0.87) \\
\hline
Observations & 17,207 & \multicolumn{2}{c|}{73} & \multicolumn{2}{c|}{422} & \multicolumn{2}{c}{14,540} & \multicolumn{2}{c|}{52} & \multicolumn{2}{c|}{1,569} & \multicolumn{2}{c}{551}\\
\hline
\end{tabular}
\end{adjustbox}
\caption{\label{evaluation_val_all} Results for REDS validation set. Standard errors reported in parentheses. \bm{$^*$}coverage = 58.42\%; \bm$^\dagger$ coverage = 58.64\%. Results in the table only represent the observations classified unambiguously.}
\end{table*}

\section{Two-way Religion Classification}\label{sec:twoway}
In this section we discuss details of a two-way religion classification with Muslim and non-Muslim classes. We follow the same pre-processing steps as before. However, we now split the data into training, validation and test sets in the ratio 80:10:10 as we have sufficiently large number of observations in the two classes. We follow the same methodology for the LR, SVM and CNN models. Additionally, we also experiment with Long short-term memory (LSTM), and CNN-LSTM architectures as described below:

\paragraph{LSTM} Recurrent Neural Networks (RNN) are designed to learn from sequential input \citep{elman1990finding}. However, as discussed by \citet{bengio1994learning}, they also suffer from vanishing/exploding gradients problem for long sequences. LSTM is a variant of RNN that handles long range dependencies by allowing the network to learn adaptively via gating mechanism \citep{hochreiter1997long}. Therefore, LSTM is widely used for NLP tasks. In our implementation, we apply a linear transformation to each hidden state output of the LSTM layer and apply max-pooling over transformed sequences to get the name encoding. The encoding is further transformed via a mapping inspired by highway layer \citep{srivastava2015training}, wherein we use a linear transformation followed by ELU activation and concatenate the transform and carry gate outputs. In our experiments, more complex architectures such as attention mechanism, stacked LSTM, bidirectional LSTM or concatenating first and last output sequence with max-pool and average-pool did not lead to improvement in validation loss.

\paragraph{CNN-LSTM} Inspired by \citet{kim2016character}, our third neural network architecture combines CNN with LSTM. We make several modifications to the model architecture. The embedding outputs are fed to separate CNN-LSTM stacks. Each such stack has a fixed CNN kernel width k, varying from 2\textendash6 and multiple filters. We use $k-1$ max pooling on CNN output. We then feed this sequence of most relevant $k$-grams encodings to an LSTM layer. We also train another LSTM layer directly from input sequence. For each LSTM layer, we perform a linear transformation over its output sequence and apply global max pooling to get k-gram name representation. Finally, we concatenate all six representations and pass them through a fully connected layer which also allows adaptive learning via our highway layer variant. We report the hyperparameter choice and the range of hyperparameter search in Table \ref{hyperparameters_twoway} and \ref{hyperparameterrange_twoway} respectively.

\begin{table}[!htbp]
\centering
\begin{adjustbox}{max width=\textwidth}
\begin{tabular}{lc}
\hline
\textbf{Parameter} & \textbf{Choice} \\ \hline
Embedding dimension & 29 (CNN); 30 (CNN concat)\\
Kernel initialization & He uniform \\
Embedding dropout & 0.1 \\
Dropout & 0.25 \\
CNN kernel sizes ($k$) & \{2, 3, 4, 5, 6\}\\
CNN filters & $\min(300, 50\cdot k + 100)$\\
CNN activation & ELU \\
LSTM hidden units & 250 \\
Highway carry bias & -2 \\
Transform activation & tanh\\
epochs & 30 \\
l2 regularization parameter & 6.95 (LR), 7.50 (LR concat); 0.21 (SVM), 0.32 (SVM concat)\\
TF-IDF Max n-grams & 5 (LR), 9 (LR concat); 9 (SVM), 10 (SVM concat) \\

\hline
\end{tabular}
\end{adjustbox}
\caption{\label{hyperparameters_twoway} Hyperparameter choice for two-way models. For the LSTM only model, we reduce LSTM hidden units to 100 with 0.15 dropout rate. For CNN only model, we also use CNN kernel size of 1.}
\end{table}

\begin{table*}[!h]
\centering
\begin{adjustbox}{max width=\textwidth}
\begin{tabular}{lc}
\hline
 & \textbf{Experimental Range} \\ \hline
Embedding dimension & 5-80\\
CNN kernel sizes ($k$) & 1\textendash7\\
CNN filters & 150\textendash300\\
CNN activation & ELU, ReLU, tanh \\
LSTM hidden units & 100-750\\
LSTM Direction & forward, reverse, bidirectional\\
LSTM pooling & first, last, average, max\\
Embedding dropout & 0\textendash0.3\\
Dropout & 0\textendash0.5\\
Highway layers & 1\textendash3\\
Transformation activation & ELU, ReLU, tanh\\
Highway layer output & add, concatenate\\
Loss & binary cross-entropy, focal-loss\\
Batch size & 32-1024\\
Optimizer & Adam, Nadam, rmsprop \\
Kernel initialization & He uniform, Glorot uniform, Glorot normal, Lecun uniform\\
l2 regularization parameter & 0\textendash100\\
TF-IDF Max n-grams & 1\textendash12\\

\hline
\end{tabular}
\end{adjustbox}
\caption{\label{hyperparameterrange_twoway} Hyperparameter range for two-way religion classification}
\end{table*}

We also experiment with a two-stage model, where we leverage the availability of two names in a household, i.e., name of the person and that of their parent/spouse in a different way. We combine probabilities $P_1$ and $P_2$ (or confidence scores in case of SVM) for these two names respectively and handcrafted features derived from them using a linear SVM model with l2 regularization. The final confidence score $C_M$ is then represented as:
\begin{equation*}
\label{stage1}
    \begin{split}
    C_M = F(f_k(P_1), f_k(P_2), g_k(P_1, P_2))
    \end{split}
\end{equation*}
 Where, $f_k$ and $g_k$ denote handcrafted features:
 \begin{equation*}
 f_k \in \{P_i, log(P_i)\} 
 \end{equation*}
 \begin{equation*}
 \begin{split}
    g_k \in  & \{P_1\cdot P_2, max(P_1, P_2), P_1\cdot log(P_2), \\ & P_2\cdot log(P_1), max(log(P_1), log(P_2))\}
 \end{split}
 \end{equation*}

We take these features to implement a non-linear decision boundary separating Muslim and non-Muslim names based on $P_1$ and $P_2$. For the final results, we use recursive feature elimination (RFE) to select only the relevant features from our feature pool. For a given feature count, the RFE algorithm chooses the set of features that contribute the most to the predictive power of the model. We then choose the \emph{optimal} number of features as those that have the highest macro-average recall on the validation set.

\begin{table*}[ht!]
\centering
\begin{adjustbox}{max width=\textwidth}
\begin{tabular}{l|c|cc|cc|c|cc|cc}
\hline
& \multicolumn{5}{c|}{\textbf{REDS}} & \multicolumn{5}{c}{\textbf{U.P. Rural Households}}\\
\hline
\multirow{2}{*}{\textbf{Models}} & \multirow{2}{*}{\bm{$F_1$}} & \multicolumn{2}{c|}{\textbf{Muslim}} & \multicolumn{2}{c|}{\textbf{Non-Muslim}} & \multirow{2}{*}{\bm{$F_1$}} & \multicolumn{2}{c|}{\textbf{Muslim}} & \multicolumn{2}{c}{\textbf{Non-Muslim}}\\
\cline{3-6} \cline{8-11}
  & & \textbf{P} & \textbf{R} & \textbf{P} & \textbf{R}& & \textbf{P} & \textbf{R} & \textbf{P} & \textbf{R}\\
 \hline
 \emph{Panel A: Single Name} & & & & & & & & &\\
Name2community\bm{$^*$} & 93.39 & 90.08 & 85.82 & 98.65 & 99.10 & 93.11 & 92.58 & 83.20 & 97.98 & 99.19 \\
Logistic Regression & 95.23 & 89.28 & 93.53& 99.35 & 98.88 & 90.24 & 79.70 & 87.01 & 97.98 & 96.60  \\
SVM & 95.64 & 90.15 & \textbf{94.10} & \textbf{99.41} & 98.97 & \textbf{91.45} & 82.70 & \textbf{87.95} & \textbf{98.13} & 97.17\\
CNN & 95.86 & \textbf{94.62} & 90.39 & 99.04 & \textbf{99.49} & 90.67 & 90.19 & 77.99 & 96.69 & \textbf{98.70} \\
LSTM & 95.89 & 92.00 & 93.05 & 99.30 & 99.19 & 90.56 & 85.62 & 81.64 & 97.20& 97.89\\
CNN-LSTM & \textbf{95.96} & 93.59 & 91.72 & 99.17 & 99.37 & 90.90 & \textbf{90.23} & 78.71 & 96.79 & 98.69\\
\hline
\emph{Panel B: Concatenated} & & & & & & & & &\\
Name2community\bm$^\dagger$ & 90.39 & 78.20 & 87.66 & 98.74 & 97.53 & 91.62 & 85.92 & 84.84 & 97.79 & 91.62 \\
Logistic Regression & 97.08 & 94.17 & 95.24 & 99.52 & 99.41 & 96.49 & 94.93 & 92.90 & 98.91 & 99.24 \\
SVM & \textbf{97.31} & 94.05 & \textbf{96.19} & \textbf{99.62} & 99.39 & \textbf{96.69} & 94.86 & 93.65 & 99.03 & 99.22 \\
CNN & \textbf{97.31} & \textbf{95.85} & 94.39 & 99.44 & \textbf{99.59} & 94.16 & \textbf{98.39} & 82.50 & 97.38 & \textbf{99.79}  \\
LSTM & 96.67 & 94.17 & 93.72 & 99.37 & 99.42 & 93.93 & 87.93 & 91.14 & 98.63 & 98.08 \\
CNN-LSTM & 97.01 & 93.90 & 95.24 & 99.52 & 99.38 & 92.85 & 81.43 & \textbf{95.16} & \textbf{99.24} & 96.67\\
\hline
\emph{Panel C: Two-Stage} & & & & & & & & &\\
Logistic Regression & 96.87 & 92.42 & 96.29 & 99.63 & 99.21 & 95.57 & 89.42 & 95.49 & 99.30 & 98.26 \\
SVM & 96.94 & 92.75 & 96.19 & 99.62 & 99.25 & \textbf{97.36} & \textbf{95.15} & 95.72 & 99.34 & \textbf{99.25} \\
CNN & \textbf{97.17} & \textbf{92.97} & 96.86 & 99.68 & \textbf{99.27} & 94.75 & 85.12 & \textbf{97.75}  & \textbf{99.65} & 97.38 \\
LSTM & 96.45 & 90.49 & 96.86 & 99.68 & 98.98 & 93.04 & 80.38 & 97.52 & 99.61 & 96.34 \\
CNN-LSTM & 96.91 & 91.89 & \textbf{97.05} & \textbf{99.70} & 99.14 & 95.02 & 86.70 & 96.73 & 99.49 & 97.72\\
\hline
Observations & 11,543 & \multicolumn{2}{c|}{1,051} & \multicolumn{2}{c|}{10,492} & 20,000 & \multicolumn{2}{c|}{2,663} & \multicolumn{2}{c}{17,337}  \\
\hline
\end{tabular}
\end{adjustbox}
\caption{\label{evaluation-test_twoway} Results for REDS and U.P. Rural Households test sets. \bm{$^*$}coverage = 65.36\% for REDS data and 57.26\% for U.P. Rural Households data; \bm$^\dagger$coverage = 75.81\% for REDS data and 74.74\% for U.P. Rural Households data. Results in the table only represent the observations classified unambiguously.}
\end{table*}

The results for the two-way classification are shown in Table \ref{evaluation-test_twoway}. Most of the comparative results remain qualitatively similar to those for the multi-religion classification. However, in panel C, which shows the results for the two-stage model, the recall for Muslim class improves substantially for the neural models. This is due to better separation between Non-Muslim and Muslim households for these models in the ($P_1$,$P_2$) space. This is an important result because of class imbalance in the data due to low Muslim proportion. Thus, neural models are now better able to classify actual Muslims and the models are less biased towards classifying a household as belonging to the majority non-Muslim class. This implies that the two-stage neural models can be expected to outperform the other models in areas with relatively high Muslim population share. \citet{chaturvedi2020importance} find that our two-stage CNN-LSTM model generalizes very well at an aggregate level using names from a census of over 25 million households in rural Uttar Pradesh. They report a correlation of 97.8\% between the Muslim household share at the sub-district (tehsil) level predicted using our model and the Muslim population share reported in the 2011 census.

Now we turn to an intuitive discussion of why we use a two-stage model with a non-linear Decision Boundary. There are a couple of ways in which we can combine an individual's name ($N_1$) with their parent/spouse's name ($N_2$) in this paper. One way is to concatenate $N_1$ and $N_2$ and increase the length of input character sequence for each observation in our data set. We can then directly calculate the collective probability $P_M$ of a household being Islamic. Alternatively, we can append $N_2$ below $N_1$ to double the number of observations in our training data. The probability $P_M$ is then calculated in two levels. In level 1, we independently obtain probabilities $P_1$ and $P_2$ on the two names using our model. Muslims in rural India do not strictly adhere to traditional Arabic names, while non-Muslims are highly unlikely to use Islamic names. Therefore, a household should be classified as Muslim if at least one name is Islamic. In other words, the algorithm should only categorize a household as non-Muslim if both the names are non-Islamic. Due to this asymmetry, we expect the decision boundary to be convex in probabilities $P_1$ and $P_2$. Since $P_1$ and $P_2$ are calculated independently, we combine these in level 2 using a non-linear decision boundary to get $P_M$. 

\begin{figure*}[!b]
    \centering
    \begin{minipage}[!]{.48\textwidth}
    \includegraphics[width=\textwidth]{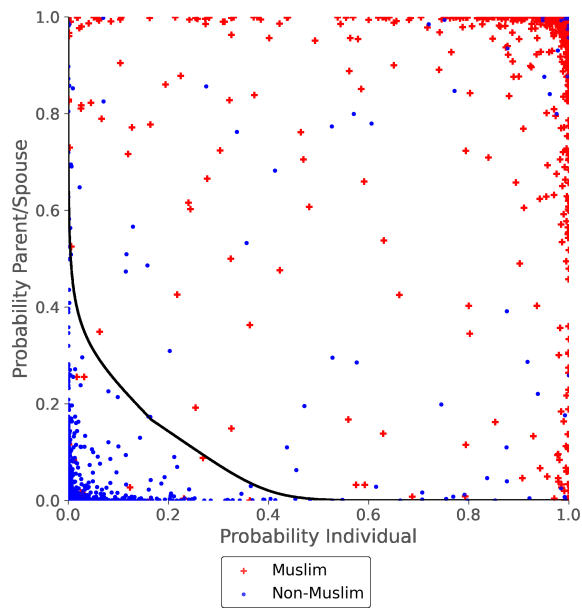}
    \end{minipage}
  \caption{SVM decision boundary using CNN model}
    \label{fig:boundary}
\end{figure*}

To illustrate, we show the second-stage decision boundary returned by SVM using first-stage probabilities from CNN model in figure \ref{fig:boundary}. We note that the second stage confers a small but statistically significant advantage at 5\% level of significance over hard coding an `or' decision boundary. This also allows us to confirm our hypothesis and derive the decision boundary in a data driven way. The households above or to the right of the boundary are classified as Muslims, whereas those below or to the left are classified as non-Muslims. For example, the household (Lal Babu, Ali Hasan) corresponding to the point (0.00, 0.99) in the figure is correctly classified as Muslim. In contrast, (Munni Rishi, Mohi Rishi) corresponding to the point (0.21, 0.09) is classified as non-Muslim. This shows that while neural models perform well at identifying character sequence patterns, they generalize better after the second stage because of the inherent uncertainty in names.

\end{document}